\documentclass{article}
\usepackage{spconf,amsmath,epsfig}
\usepackage{caption}
\usepackage{subcaption}
\usepackage{multirow}
\usepackage{multicol}
\usepackage{url}

\title{Multimodal Object Detection in Remote Sensing}
\name{A. Belmouhcine, J. C. Burnel, L. Courtrai, M. T. Pham, S. Lefèvre}
\address{IRISA, Université Bretagne Sud, UMR 6074, 56000 Vannes, France}
\begin{document}
%
\maketitle
\begin{abstract}
Object detection in remote sensing is a crucial computer vision task that has seen significant advancements with deep learning techniques. However, most existing works in this area focus on the use of generic object detection and do not leverage the potential of multimodal data fusion. In this paper, we present a comparison of methods for multimodal object detection in remote sensing, survey available multimodal datasets suitable for evaluation, and discuss future directions.
\end{abstract}
\section{Introduction}

 Object detection in remote sensing has seen significant advancements in recent years, thanks to deep learning techniques~\cite{li2020object,li2022deep}. However, remote sensing images pose unique challenges due to their distinct structure and specific requirements. Objects in these images are often small and cover only a few pixels due to the high scale of the images. Additionally, objects exhibit arbitrary perspective transformations, necessitating the use of oriented bounding boxes~\cite{li2020object,li2022deep}. Multimodal sensing, which involves the fusion of multiple data sources
 , is crucial for object detection. However, most existing works in remote sensing object detection focus solely on generic object detection and do not leverage the multimodal aspect of the images~\cite{li2020object,li2022deep}.

Data fusion plays a vital role in remote sensing tasks~\cite{LI2022102926}, combining information from multiple modalities to obtain more accurate and informative results. While data fusion has been extensively applied in various Earth Observation applications, such as pansharpening, classification, and semantic segmentation~\cite{LI2022102926,hong2021overview}, its application to object detection~\cite{zhang2023superyolo, 9273212, QINGYUN2022108786}  has been largely unexplored. Object detection could greatly benefit from multimodal fusion, as the potential of multimodal imagery has been demonstrated in computer vision, where combining RGB with NIR has improved object detection, particularly in bad weather conditions~\cite{zhang2020multispectral}. While public multimodal datasets like FLIR~\cite{zhang2020multispectral}, KAIST~\cite{kaist}, and LLVIP~\cite{llvip} exist, there is a need to transfer these developments to aerial and satellite remote sensing data.

The limited availability of multimodal datasets and labeled training data is a major challenge in multimodal object detection for remote sensing~\cite{razakarivony2016vehicle}. However, recent efforts have introduced the GeoImageNet dataset~\cite{li2022geoimagenet}. Moreover, semantic segmentation datasets, e.g., ISPRS 2D Semantic Labeling Contest -- Potsdam dataset~\footnote{\url{https://www.isprs.org/education/benchmarks/UrbanSemLab/2d-sem-label-potsdam.aspx}}, can be converted into object detection datasets \cite{audebert2017segment}. Finally, SeaDronesSee~\cite{varga2022seadronessee} is a unique dataset that focuses on object detection and tracking of people in open water, aiming to bridge the gap between land-based and sea-based vision systems. 


This paper gives an overview of available multimodal object detection datasets, presents an experimental comparison of remote sensing multimodal object detectors on those datasets, discusses results, and provides recommendations.

\section{Datasets}

\begin{figure*}
    \centering
    
    \includegraphics[width=.15\textwidth, height=.15\textwidth]{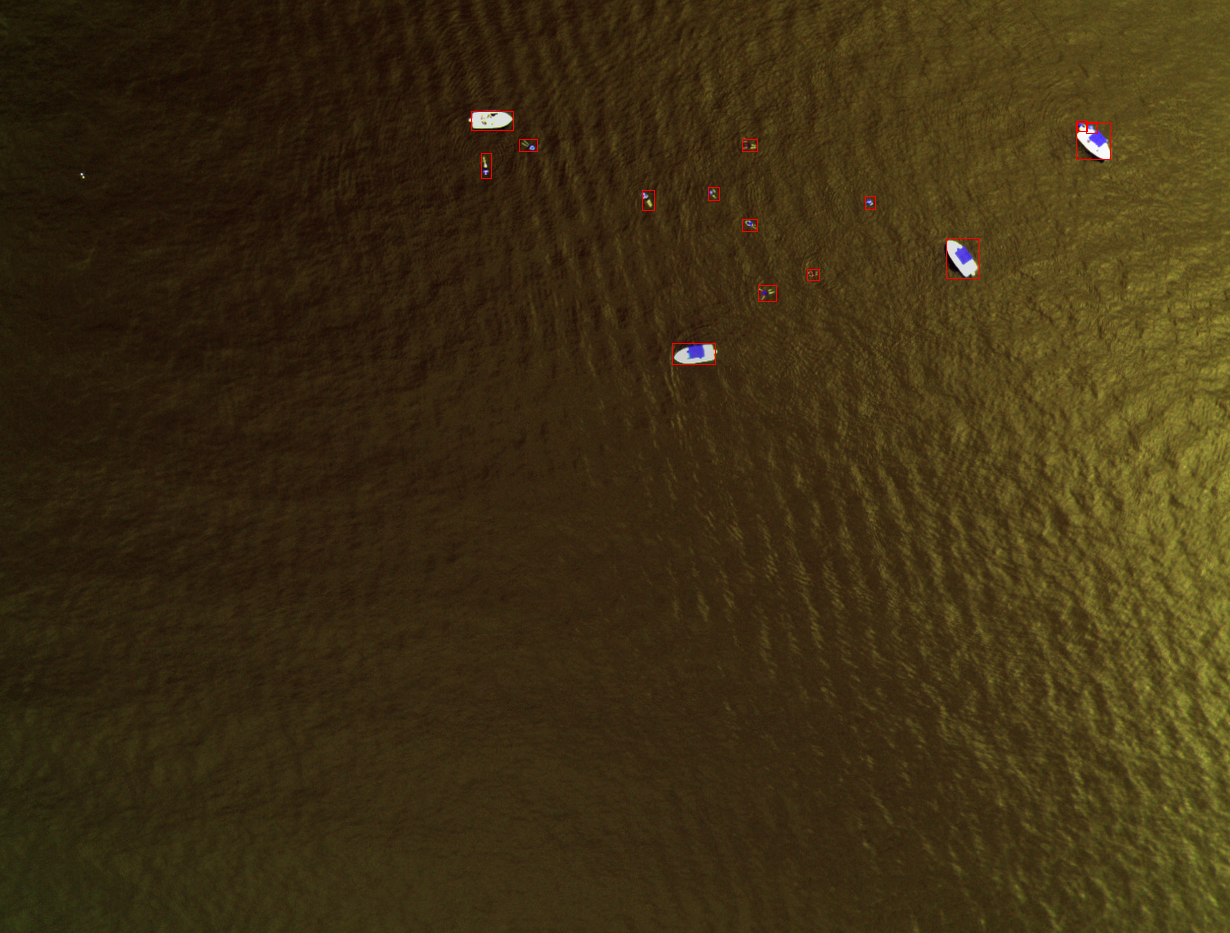}
    \includegraphics[width=.15\textwidth, height=.15\textwidth]{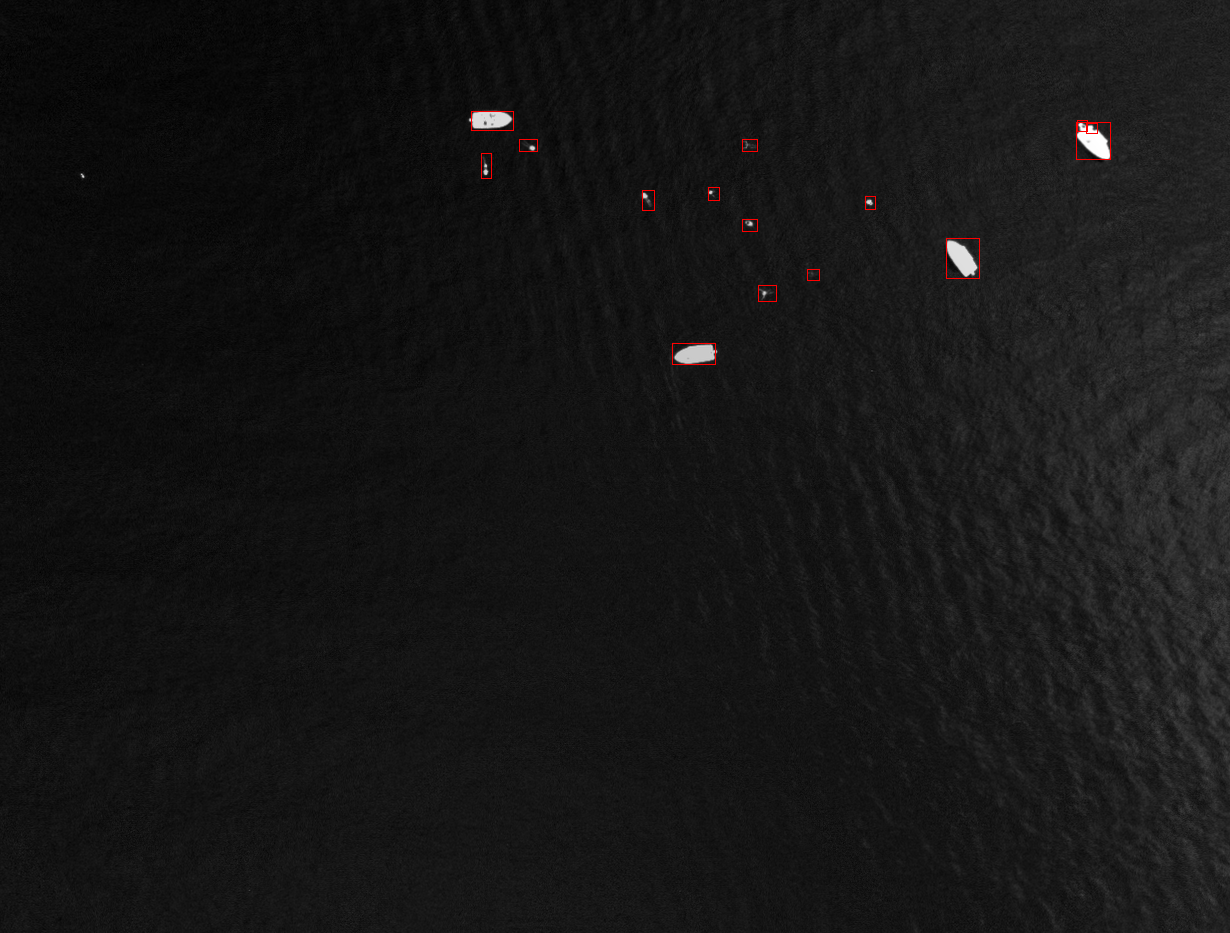}
    \includegraphics[width=.15\textwidth, height=.15\textwidth]{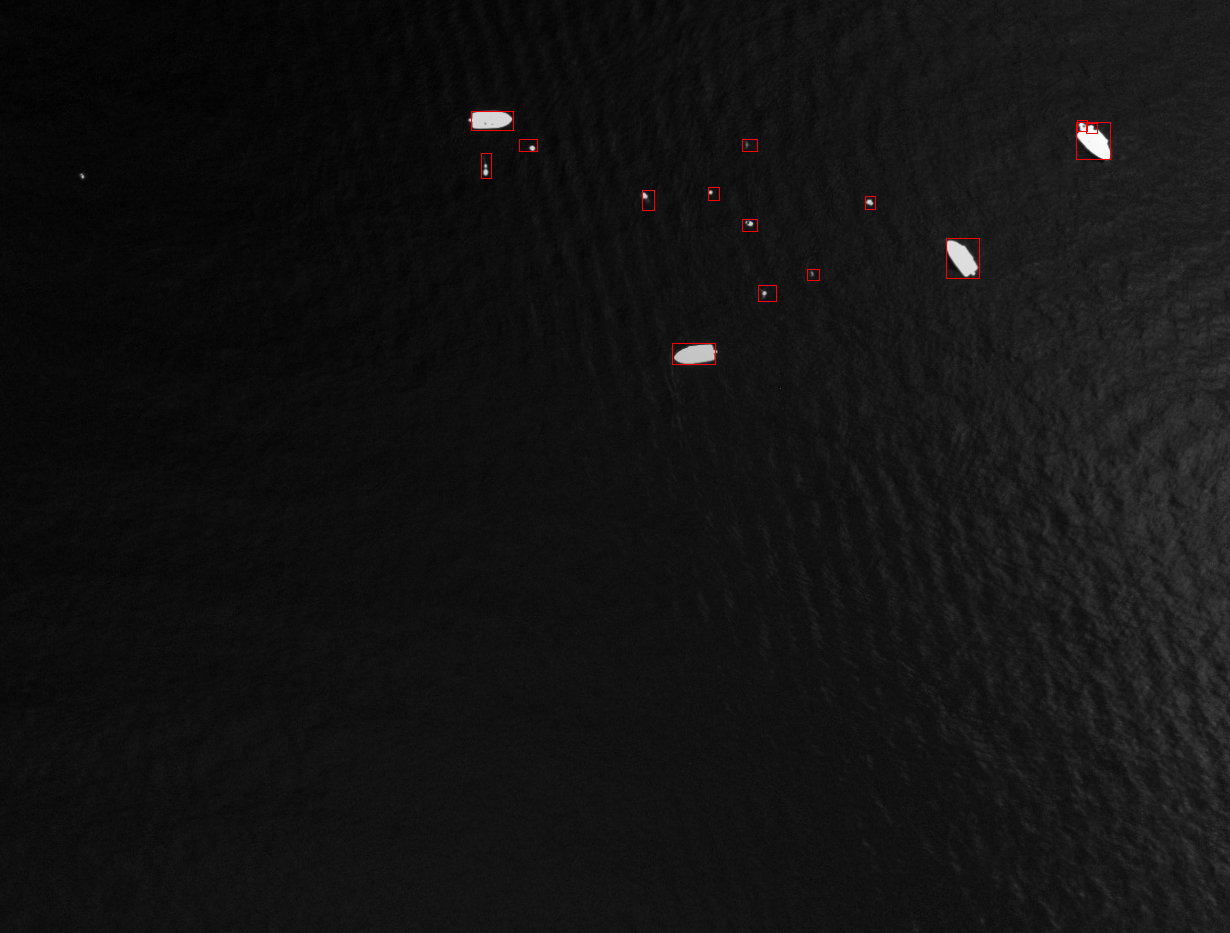}
    \hfill
    \includegraphics[width=.15\textwidth, height=.15\textwidth]{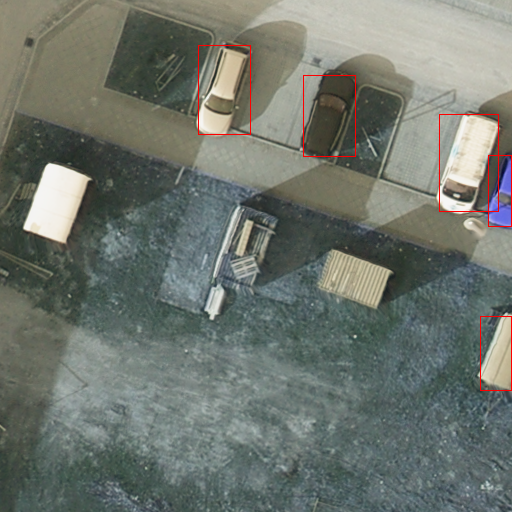}
    \includegraphics[width=.15\textwidth, height=.15\textwidth]{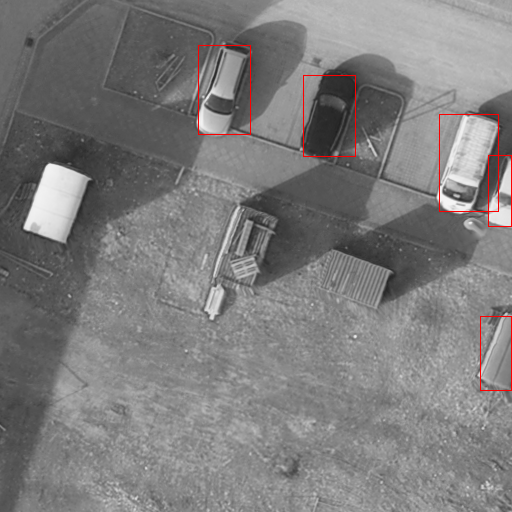} 
    \includegraphics[width=.15\textwidth, height=.15\textwidth]{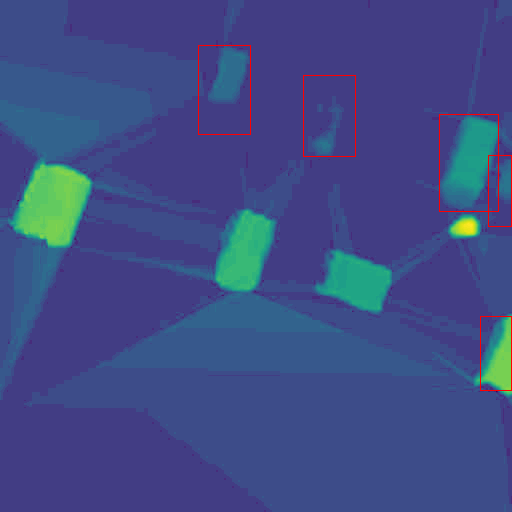}
    \\[1em]
    \hfill
    \includegraphics[width=.15\textwidth, height=.15\textwidth]
    {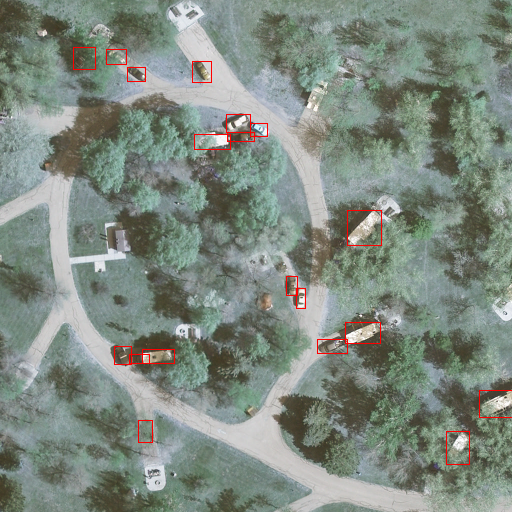}
    \includegraphics[width=.15\textwidth, height=.15\textwidth]{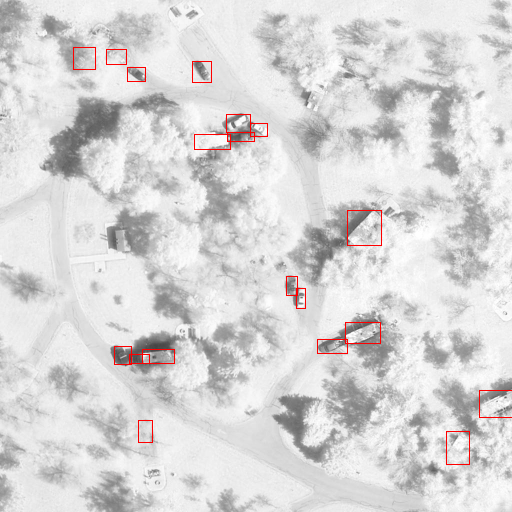}
    \hfill\hfill
    \includegraphics[width=.15\textwidth, height=.15\textwidth]
    {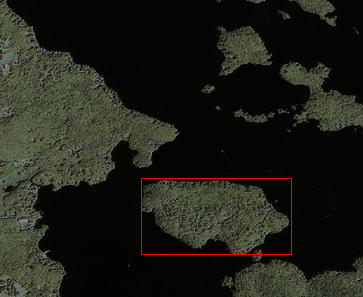}
    \includegraphics[width=.15\textwidth, height=.15\textwidth]{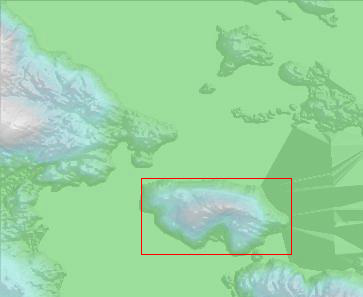}
    \hfill~
  \caption{Multimodal images from 4 different datasets. From left to right, top to bottom, samples from: SeaDronesSee (RGB, NIR, RedEdge), ISPRS-Potsdam (RGB, NIR, DSM), VEDAI (RGB, NIR), and GeoImageNet (RGB, DEM). Red bounding boxes surround all objects of interest.}
  \label{fig:datasets}
\end{figure*}

Several unimodal datasets have been proposed in remote sensing, e.g., DIOR~\footnote{\url{https://universe.roboflow.com/new-workspace-ghppr/dior-dataset-riv6b}}, DOTA~\cite{9560031}, and NWPU VHR-10~\cite{su2019object}. 
However, relying solely on RGB remote sensing images is susceptible to poor lighting, occlusion, and adverse weather conditions. Additionally, we have identified the presence of multimodal remote sensing datasets, such as VEDAI, GeoImageNet, SeaDronesSee, and ISPRS Potsdam, which offer a wider range of data sources for analysis and testing of object detectors. Figure~\ref{fig:datasets} shows examples of images for all datasets.

\subsection{ISPRS Potsdam}
The ISPRS Potsdam Semantic Labeling dataset\textsuperscript{1} comprises 38 ortho-rectified aerial images with a spatial resolution of 5 cm per pixel taken over Potsdam, Germany. This dataset has multiple modalities, notably orthophotos (RGB, IR) and Digital Surface Model (DSM). We used the already computed normalized DSM for this dataset. Regarding tiles splitting, 10 were used for training, 18 were kept without any label, five were used for validation, and five were left for testing.
To prepare the dataset for object detection, we took the segmentation masks of all cars converted to non-oriented bounding boxes, as already done in \cite{audebert2017segment}. Then, we slid a nonoverlapping 512-sized window over each tile and kept only patches containing objects. This results in 540 patches for training, 240 for validation, and 339 for testing.

\subsection{VEDAI}
The VEDAI dataset~\cite{razakarivony2016vehicle} is a subset of the Utah Automated Geographic Reference Center (AGRC) dataset. AGRC images have a resolution of about 12.5 $\times$ 12.5 cm per pixel. The VEDAI dataset comprises 1,246 smaller images cropped from AGRC, available in two resolutions: 1,024 $\times$ 1,024 and 512 $\times$ 512. These images were carefully chosen to include diverse backgrounds like grass, highways, mountains, and urban areas. The VEDAI dataset focuses on 11 different vehicle classes, but, as was done in~\cite{9273212}, we excluded classes with less than 50 instances: planes, motorcycles, and buses.
For our training and testing purposes, we utilized 1,089 images for training and 121 images for testing.
We used non-oriented bounding boxes to ensure consistency and facilitate comparisons in our evaluation.

\subsection{GeoImageNet}

GeoImageNet~\cite{li2022geoimagenet} is a benchmark dataset specifically designed for GeoAI applications, focusing on natural features. It incorporates multi-source data, including RGB remote sensing images and Digital Elevation Model (DEM), which provides valuable spatial context for object detection and classification tasks. 
The dataset consists of 876 image pairs belonging to 6 natural feature categories: basin, bay, island, lake, ridge, and valley. The spatial resolution of RGB images and DEM is 10 meters. The image sizes vary, ranging from 283 $\times$ 213 to 4,584 $\times$ 4,401, with an average size of 635.44 $\times$ 593.45. We kept the same data splitting done by the dataset providers. We used 698 pairs of images for training and 178 for testing.  

\subsection{SeaDronesSee}
SeaDronesSee dataset~\cite{varga2022seadronessee} disrupts the current trend of multimodal datasets that primarily focus on land-based traffic scenarios by offering a unique dataset centered around people in open water environments. This dataset is a significant benchmark for object detection and tracking, bridging the gap between vision systems designed for land and those tailored explicitly for sea settings. The dataset includes multiple modalities, namely RGB, NIR, and RedEdge, providing comprehensive visual information. In this paper, the object categories in this dataset are simplified into two supercategories: person (including subcategories such as swimmer, floater, west person, and human) and boat. The data are captured using an Unmanned Aerial Vehicle (UAV) equipped with a camera, ensuring that diverse images are taken from various altitudes (ranging from 5 to 260 meters) and viewing angles (ranging from 0 to 90 degrees). 
SeaDronesSee comprises 307 multispectral images, with 242 images dedicated to training, as specified by the authors~\cite{varga2022seadronessee}.

\begin{table*}[t!]
        \centering
         \resizebox{\textwidth}{!}{
        \begin{tabular}{|c|c|cc|cc|cc|}
        \hline
        &&\multicolumn{4}{c|}{\textbf{Supervised}}&\multicolumn{2}{c|}{\textbf{Semi-supervised}}\\
        \cline{3-8}
        \textbf{Dataset}&\textbf{Modality}&\multicolumn{2}{c|}{\textbf{YOLOrs}}&\multicolumn{2}{c|}{\textbf{SuperYOLO}}&\multicolumn{2}{c|}{\textbf{LLMM}}\\
        \cline{3-8}        
         &&\textbf{mAP@50 $\uparrow$}&\textbf{mAP@0.5:0.95 $ \uparrow$}&\textbf{mAP@50 $\uparrow$}&\textbf{mAP@0.5:0.95 $ \uparrow$}&\textbf{mAP@50 $\uparrow$}&\textbf{mAP@0.5:0.95 $ \uparrow$}\\
         \hline
        &RGB&0.655&0.394& 0.630&0.366&--&--\\
        VEDAI&IR&0.608&0.367& 0.599&0.358&--&--\\
        &RGB+IR&0.531&0.306&\textbf{0.688}&\textbf{0.407}&0.055&0.0167\\
         \hline
        &RGB&0.931&0.615&0.939&\textbf{0.632}&--&--\\
        SeaDronesSee&IR&0.885&0.552&0.915&0.578&--&--\\
        &RGB+IR&0.907&0.578&\textbf{0.944}&0.619&0.123&0.023\\
        
        \hline
        &RGB&0.820&0.581&0.841&0.586&--&--\\
        ISPRS Potsdam&DSM&0.251&0.085&0.254&0.079&--&--\\
        &RGB+DSM&0.809&0.559&\textbf{0.849}&\textbf{0.595}&0.793&0.363\\

        \hline
        &RGB&0.420&0.224&0.324&0.150&--&--\\
        GeoImageNet&DEM&\textbf{0.533}&\textbf{0.269}&0.426&0.167&--&--\\
        &RGB+DEM& 0.134&0.068&0.323&0.144&0.188&0.071\\

        \hline
        \end{tabular}
        }
        \caption{Detection performance comparison of YOLOrs, SuperYOLO and LLMM on four datasets with single and multi modalities}
        \label{tab:res}
\end{table*}

\section{MULTIMODAL DETECTORS}

The review by Zheng et al.~\cite{li2022deep} highlights the importance of leveraging the multimodal aspect of remote sensing images in object detection. While most existing works focus on general object detection, they fail to take advantage of the various modalities available in remote sensing, such as RGB, SAR, LiDAR, IR, PAN, and Multispectral. Integrating multimodal data can enhance the robustness and reliability of object detectors. Data fusion is crucial in dealing with multimodal remote sensing data, combining information from multiple modalities to achieve more accurate results. However, the application of multimodal data fusion to object detection in remote sensing remains limited. Few fusion approaches have been proposed, including early fusion, middle fusion, late fusion, and decision fusion, with middle fusion showing promising performance. Jiaxin et al.~\cite{LI2022102926} emphasize that object detection should receive more attention in multimodal fusion tasks, as it is a significant aspect of remote sensing multimodal data analysis.

In the field of multimodal object detection in remote sensing, two specific algorithms have been proposed: Super\-YOLO by Zhang et al.~\cite{zhang2023superyolo} and YOLOrs by Sharma et al.~\cite{9273212}. SuperYOLO utilizes an encoder-decoder super-resolution module for a pixel-level fusion of RGB and IR images. It focuses on detecting small objects within vast background areas and achieves high-resolution features through fusion. YOLOrs, on the other hand, targets multimodal remote sensing images and provides oriented bounding boxes. It applies six YOLOv5 layers on each modality and fuses them using concatenation or cross-product of feature maps. Both algorithms demonstrate the benefits of multimodal fusion for object detection in remote sensing on the VEDAI dataset.

The attention mechanism has improved many computer vision algorithms for remote sensing images~\cite{ghaffarian2021effect}. Fang and Wang~\cite{QINGYUN2022108786} proposed a Cross-Modality Attentive Feature Fusion (CMAFF) and combined it with YOLOv5s to obtain YOLOFusion. It comprises two modules: the Differential Enhancive Module and Common Selective Module, and aims to enhance modality-specific features while cherry-picking shared features to avoid redundancy. 


Regarding semi-supervision, some algorithms proposed using labels for a subset of the available data where only one modality is available to add other modalities without the help of any other labels. In \cite{burnel2022ajout}, a solution called ``Less Labels More Modalities'' (aka LLMM) was originally designed for semantic segmentation on ISPRS Potsdam, but the method also applies to object detection. Their method uses self-training to produce pseudo-labels and an auxiliary network called contributor to generate a modality.

\section{Experiments}

SuperYOLO and YOLOrs are specialized models specifically designed for detecting small objects while disregarding larger ones, as evident from their performances on the GeoImageNet dataset shown in Table~\ref{tab:res}. Since we focus on multimodal data fusion for object detection in remote sensing, we do not consider the super-resolution module. It is crucial to ensure a fair comparison between SuperYOLO and YOLOrs, as they were applied to different datasets. The comparison done in \cite{zhang2023superyolo} was not fair. Indeed, YOLOrs was applied to oriented VEDAI, while SuperYOLO was applied to non-oriented VEDAI.
Moreover, the comparison, contained in \cite{zhang2023superyolo}, between SuperYOLO's mAP@50 results and YOLOrs' mAP@0.1 results is inappropriate. Thus, we decided to rerun YOLOrs and SuperYOLO on four datasets using the same settings to address these issues. Our analysis reveals that YOLOrs' simple fusion approach (concatenation) is unreliable, as it fails to improve results in any of the cases presented in Table~\ref{tab:res}. To accurately evaluate the models' performance, we calculated the mAP using a low confidence threshold, as it should be, instead of using a high confidence threshold as done in \cite{9273212}. Besides, results in Table~\ref{tab:res} demonstrate the strong performance of SuperYOLO with its compact multimodal fusion module across all datasets except GeoImageNet, because YOLOrs and SuperYOLO are specialized in small objects. However, remote sensing objects are not always small~\cite{li2022geoimagenet}. Furthermore, those algorithms do not learn any weighting schema for different modalities, which explains the deterioration of performances in YOLOrs and the slight improvement of results with SuperYOLO.  YOLOFusion processes different features differently; the modality-specific features are retained and enhanced, while the modality-shared features are cherry-picked from the RGB and IR modalities. Unfortunately, the code of YOLOFusion is no longer available in the authors' Git repository~\footnote{\url{https://github.com/DocF/CMAFF}}. Thus, we were not able to perform any experimental assessment of this model.

Regarding the LLMM method, results could be better because the approach uses pseudo-labels obtained by a single modality teacher to train a bimodality student, and the lack of training data leads to a model that overfits on the single modality and fails to transfer to both modalities. 

\section{conclusion}

In this paper, we conducted a brief review of multimodal object detection in remote sensing. SuperYOLO outperformed YOLOrs in detecting small objects in remote sensing data. YOLOrs' fusion approach was ineffective, while Super\-YOLO's multimodal fusion module consistently demonstrated exemplary performance. Future research can explore enhancements in fusion techniques to improve object detection in remote sensing, especially for detecting small and large objects. Moreover, many multimodal object detectors have been proposed for natural images. Yet, their effectiveness remains to be tested on remote sensing data.
\bibliographystyle{IEEEbib}
\bibliography{refs}

\end{document}